\documentclass[letterpaper, 10 pt, conference]{ieeeconf}  

\IEEEoverridecommandlockouts                              

\usepackage{bm}
\usepackage{graphicx}
\usepackage{cite}
\usepackage{amsmath,amssymb,amsfonts}
\usepackage{algorithm}
\usepackage{algorithmic}
\usepackage{comment}
\usepackage{caption}
\usepackage{subcaption}
\usepackage{multirow}
\usepackage{float}    
\usepackage{placeins} 
\usepackage{hyperref}
\usepackage{booktabs}
\usepackage{svg}
\usepackage{makecell}
\usepackage{tabularx}

\usepackage[acronym]{glossaries}

\newacronym{adas}{ADAS}{Advanced Driver-Assistance Systems}
\newacronym{ads}{ADS}{Automated Driving Systems}
\newacronym{cnn}{CNN}{Convolutional Neural Network}
\newacronym{knn}{KNN}{K-Nearest Neighbors}
\newacronym{ldw}{LDW}{Lane Departure Warning}
\newacronym{lka}{LKA}{Lane Keeping Assist}
\newacronym{mae}{MAE}{Mean Absolute Error}
\newacronym{mut}{MuT}{Model under Test}
\newacronym{nms}{NMS}{Non Maximum Suppression}
\newacronym{ood}{OOD}{Out-of-Distribution}
\newacronym{tp}{TP}{True Positive}
\newacronym{tpr}{TPR}{True Positive Rate}
\newacronym{laneperf}{LanePerf}{Lane Performance Estimation Framewok}
\newacronym{ac}{AC}{Average Confidence}
\newacronym{atc}{ATC}{Average Threshold Confidence}
\newacronym{doc}{DOC}{Difference of Confidence}
\newacronym{fid}{FID}{Frechet Inception Distance}
\newacronym{energy}{EBM}{Energy-based Model}
\newacronym{fcn}{FCN}{Fully Connected Network}
\newacronym{rho}{Spearman’s $\rho$}{Spearman’s rank correlation coefficient}
\newacronym{iou}{IoU}{Intersection over Union}

\title{\LARGE \bf LanePerf: a Performance Estimation Framework for Lane Detection}

\author{Yin Wu$^{1, 2}$, Daniel Slieter$^{1}$, Ahmed Abouelazm$^{3}$, \\ Christian Hubschneider$^{2, 3}$, and J. Marius Zöllner$^{2, 3}$
\thanks{$^{1}$Authors are with the CARIAD SE, Germany \newline {\tt\small yin.wu@cariad.technology}}
\thanks{$^{2}$Authors are with the Karlsruhe Institute of Technology, Germany}%
\thanks{$^{3}$Authors are with the FZI Research Center for Information Technology, Germany}%
}

\begin{document}
\maketitle
\thispagestyle{empty}
\pagestyle{empty}

\begin{abstract}
Lane detection is a critical component of \gls{adas} and \gls{ads}, providing essential spatial information for lateral control. However, domain shifts often undermine model reliability when deployed in new environments. Ensuring the robustness and safety of lane detection models typically requires collecting and annotating target domain data, which is resource-intensive. Estimating model performance without ground-truth labels offers a promising alternative for efficient robustness assessment, yet remains underexplored in lane detection. While previous work has addressed performance estimation in image classification, these methods are not directly applicable to lane detection tasks. This paper first adapts five well-performing performance estimation methods from image classification to lane detection, building a baseline. Addressing the limitations of prior approaches that solely rely on softmax scores or lane features, we further propose a new \gls{laneperf}, which integrates image and lane features using a pretrained image encoder and a DeepSets-based architecture, effectively handling zero-lane detection scenarios and large domain-shift cases. Extensive experiments on the OpenLane dataset, covering diverse domain shifts (scenes, weather, hours), demonstrate that our \gls{laneperf} outperforms all baselines, achieving a lower \gls{mae} of 0.117 and a higher \gls{rho} of 0.727. These findings pave the way for robust, label-free performance estimation in \gls{adas}, supporting more efficient testing and improved safety in challenging driving scenarios.


    
\end{abstract}
\section{Introduction}
\label{sec:Introduction}

\begin{figure}[t]
    \centering
    \includegraphics[width=\linewidth]{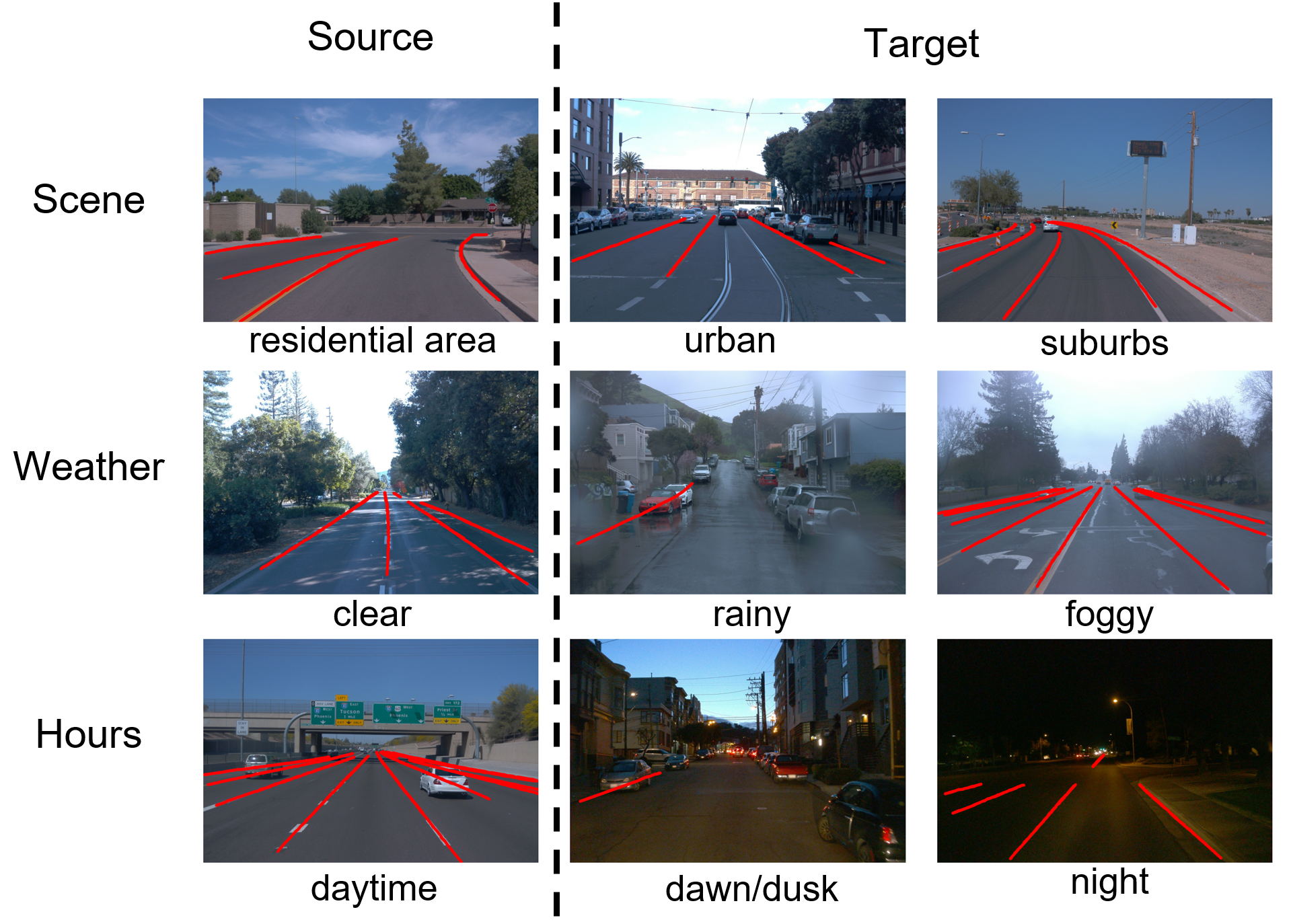}
    \caption{Example images of domain shifts in Scene, Weather, and Hours. The left column shows examples from the source domain, while the right two columns present examples from the target domains.}
    \label{fig:images}
\end{figure}

Lane detection plays a crucial role in \gls{adas} and \gls{ads}, serving as the foundation for functionalities such as \gls{lka} and \gls{ldw}. With the rapid advancements in deep learning, recent lane detection methods have achieved state-of-the-art performance on closed datasets~\cite{lanenet2018, scnn2018, lineCNN2020, laneatt2021, polylanenet2021, clrnet2022}. However, despite their success in a scene from training distribution, deploying these models in real-world scenarios remains challenging due to domain shift.
Domain shift refers to the performance degradation of machine learning models when deployed on data distributions different from those seen data during training~\cite{domainshift2009}. For instance, as shown in Figure~\ref{fig:images}, a model trained on well-lit daytime data may fail when exposed to nighttime darkness or glare during dawn or dusk. Such domain discrepancies between the source and target environments can render \gls{adas} functions unreliable and compromise the safety of \gls{ads} during deployment. Therefore, it is crucial to evaluate model performance in unseen target domains before releasing systems to the market.

A straightforward but costly solution is to conduct extensive test drives in the target domain. This involves prototype preparation, test execution, data collection and handling, annotation, and performance evaluation. However, this process is highly resource-intensive and time-consuming. Among these steps, data annotation and evaluation are particularly expensive, and become impractical when dealing with large-scale data—especially the continuously collected fleet data, which is generated daily by numerous vehicles in diverse real-world conditions.


To address these challenges, Performance Estimation—also known as AutoEval—has been proposed to assess model performance without requiring ground-truth labels~\cite{frechet2021}. Instead of calculating traditional metrics by comparing predictions to annotated labels, performance estimation methods infer the performance of \gls{mut} by leveraging internal model features.
Most prior work focuses on estimating image classification accuracy and can be categorized into three types based on their core principles: confidence-based~\cite{baselineood2016, atc2022, doc2021}, distance-based~\cite{maha2018, frechet2021, knn2022}and density-based methods~\cite{energy2021, energy2022}.

While these methods have shown promising results in image classification, adapting them to lane detection remains largely underexplored. This is because lane detection poses fundamentally different challenges compared to image classification, such as spatially structured outputs, variable numbers of detected lanes, and sensitivity to environmental context, making direct transfer of existing techniques non-trivial.



In this work, we first adapt five well-performing performance estimation methods from image classification to the lane detection task. We analyze the limitations of existing baselines in handling lane detection tasks and propose a novel framework that addresses key challenges in lane detection. In particular, our approach effectively handles zero-output scenarios and complex domain shifts. We conduct comprehensive experiments addressing three kinds of domain shift on the OpenLane dataset. Our results show that the proposed method achieves lower \gls{mae} and higher \gls{rho} in predicting the F1 score, significantly outperforming existing baselines

\section{Related Work}  
\label{sec:related_works}


\subsection{Lane detection.}
Based on lane representation, lane detection models can be grouped into segmentation-based, parameter-based, and anchor-based methods.

\subsubsection{Segmentation-based methods}
these methods treat lane detection as semantic segmentation task. Early work, such as LaneNet~\cite{lanenet2018}, follows a two-stage structure, where lanes are first segmented at the pixel level and then clustered into lane instances. SCNN~\cite{scnn2018} uses a top-down one-stage approach,  where each lane is treated as a separate category. However, since pixel-level prediction usually demands high resolution, these methods often suffer high computation costs, making them less efficient for real-time applications.

\subsubsection{Parameter-based methods}
PolyLaneNet~\cite{polylanenet2021}, represent lanes using curve equations rather than discrete coordinates. Compared to pixel-level representations, these methods require significantly fewer parameters to regress and generally offer faster inference speeds. However, their compact representation also makes them more sensitive to domain shifts, as even small environmental variations can lead to noticeable prediction errors.
\subsubsection{Anchor-based methods}:
anchor-based methods are widely adopted in object detection tasks and have also gained popularity in lane detection. These methods use a large number of predefined lines as anchors and learn the horizontal offsets relative to them. The predicted anchor outputs are typically filtered using a non-maximum suppression (\gls{nms}) to eliminate redundant detections. Recent studies~\cite{lanedetsurvey2024, lanedettranssurvey2025} have shown that anchor-based methods offer a favorable trade-off between accuracy and inference speed. One representative method is CLRNet~\cite{clrnet2022}, which employs a multi-scale feature pyramid and a ROIGather module to enhance both accuracy and efficiency, achieving top-ranking results on multiple benchmarks such as CULane, LLAMAS, and TuSimple~\cite{scnn2018, llamas2019, tusimple2020}. 

\begin{figure}[t]
    \centering
    \includegraphics[width=\linewidth]{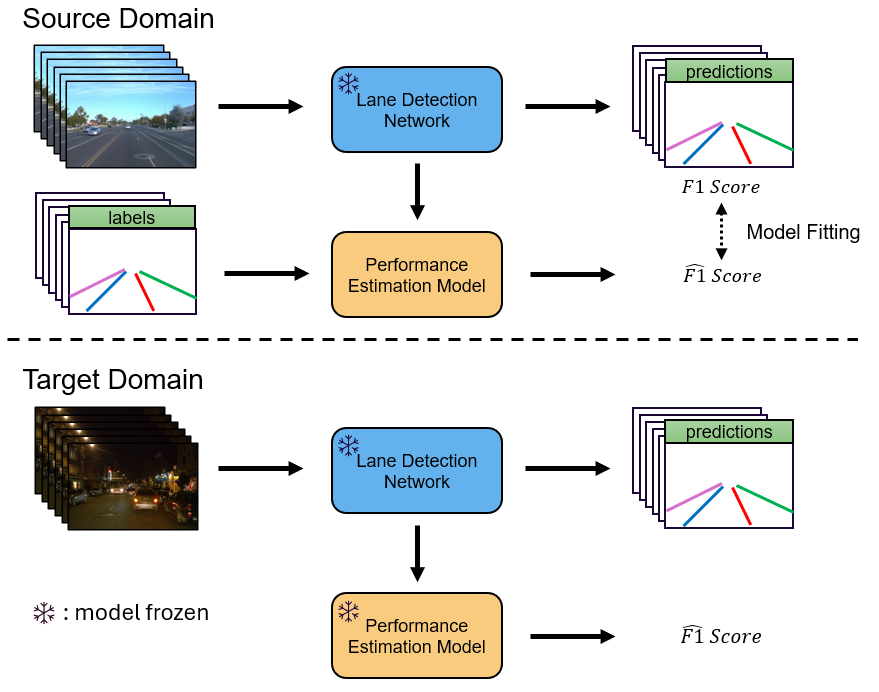}
    \caption{The lane detection model is trained on the source domain training dataset and subsequently frozen. A performance estimation model is calibrated on the source domain validation dataset, supervised by ground-truth labels to predict the F1 score. Both the lane detection model and the calibrated performance estimation model are then jointly deployed on the target domain to enable label-free performance estimation.}
    \label{fig:structure}
\end{figure}

In this study, we choose CLRNet as our \gls{mut}, due to its strong performance and increasing attention in the lane detection community.

\subsection{Domain Adaptation for Lane Detection}

Despite domain shift being a critical challenge for lane detection models, few studies have addressed this issue. Some studies tackle it from a domain adaptation perspective. For example, Wen et al.~\cite{bridelanedetDA2020} proposed unifying viewpoints across different datasets by estimating a transformation matrix. Zhou et al.~\cite{weaklyDA2024} introduced a weakly supervised method to adapt the model using only lane numbers. While these methods aim to improve model generalization across domains, they do not address the critical problem of estimating model performance under domain shift. However, evaluating the robustness of lane detection models, especially label-free methods, remains an open and underexplored challenge.



\subsection{Performance Estimation in Vision Tasks.} Performance estimation aims to predict model accuracy on unlabeled data, particularly when the target distribution differs from the training distribution. Most existing research has focused on image classification tasks, and proposed methods can be broadly categorized into four groups: confidence-based~\cite{tempscaling2017, doc2021, atc2022}, distance-based~\cite{maha2018, frechet2021}, density-based~\cite{energy2021, energy2022}, and agreement-based methods~\cite{agreement2022}. 
\subsubsection{Confidence-based methods} rely on the softmax scores of the predicted class labels. The simplest approach calculates the average confidence of predictions~\cite{baselineood2016}. However, such methods are often reported to produce overconfident estimates (softmax score). To mitigate this, a temperature scaling factor, fitted using validation datasets, is typically applied to calibrate the predicted confidence scores to reflect true accuracy more reliably~\cite{tempscaling2017}. Another kind of approach computes an offset or threshold with validation datasets of the same domain. \gls{doc} computes a confidence offset by measuring the difference between model confidence scores on the training and validation datasets, subsequently applying this bias to estimate performance on target data~\cite{doc2021}. \glspl{atc}~\cite{atc2022} adopts a threshold-based strategy, learning a confidence threshold and predicting accuracy as the proportion of samples whose confidence exceeds this threshold. 
\subsubsection{Distance-based methods} assume that feature vectors belonging to the same class should cluster closely together, whereas outliers or samples from different distributions appear as distant points. Deng et al.~\cite{frechet2021} utilize the \gls{fid} to measure distributional discrepancy between source and target domains under the assumption of Gaussian feature distributions. As the distribution assumption may not hold, Sun et al.~\cite{knn2022} use \gls{knn} to identify out-of-distribution samples from the target domain.
\subsubsection{Energy-based methods} estimate model confidence by computing an energy score from the output logits, where lower energy indicates higher model certainty~\cite{energy2021, perfpred_recon_seg2022}. Compared to softmax-based methods, energy scores are less susceptible to the overconfidence issue.

\subsubsection{Agreement-based methods} rely on multiple independently trained models and assess performance based on the consistency of their predictions~\cite{agreement2022}. Although these methods have shown promising results, their high computational cost limits practical applicability. Therefore, they are not considered in this study.

Beyond image classification, recent works have also explored performance estimation in semantic segmentation tasks~\cite{perfpred_seg2021, perfpred_recon_seg2022}, typically employing auxiliary models trained on intermediate features extracted from the main \gls{mut}. However, compared to lane detection tasks, semantic segmentation produces a consistently structured pixel-wise output. In contrast, lane detection outputs vary considerably, including scenarios with no lanes or numerous lane instances, presenting additional complexity for performance estimation.

In this work, we adapt five representative methods based on three underlying principles as our baselines: \gls{ac}, \gls{atc}, \gls{doc}, \gls{fid}, and \gls{energy}. The detailed adaptations and formulations are presented in Section~\ref{sec:methods}.

\section{Performance Estimation Baselines}
\label{sec:methods}


This section first introduces the preliminary definitions of the performance estimation task for lane detection. It then presents the underlying principles and formulations of five baseline methods, originally proposed for image classification.

\subsection{Problem Definition}

Let \( f \) denote a lane detection model, defined as \( f: x_i \rightarrow \hat{y}_i \), where \( x_i \) represents an input image and \( \hat{y}_i \) denotes the predicted set of lane instances for \( x_i \). The model \( f \) is trained on a labeled source domain dataset \( D^s = \{(x_i, y_i)\}_{i=1}^{M} \), where \( y_i = \{l_{i1}, l_{i2}, \ldots\} \) denotes the ground-truth annotations of the input image $x_i$, and each \( l_{ij} \) represents an individual lane instance in the image. Each \( y_i \) and \( \hat{y}_i \) is a set that can be empty when no lanes are present in the scene.

Given an unlabeled target domain dataset \( D^t = \{x_i\}_{i=1}^{N} \), the objective is to estimate the performance metric of \( f \) on \( D^t \). Additionally, a collection of validation datasets \( \{D^{v,(k)}\}_{k=1}^{K} \), that are drawn from the same distribution as \( D^s \) is available. They are not used during the training of \( f \), but are utilized to fit the performance estimation model \( E \).
The estimated score on \( D^t \) is then given by:

\[
Score = E(f, D^t).
\]

We adopt the F1 score as the performance metric in this work, as it is the most commonly used and effectively balances precision and recall, both of which are critical for lane detection tasks. Alternative metrics such as 
$mF1$ from~\cite{clrnet2022} and point-wise matching from~\cite{tusimple2020} introduce different IoU thresholds or lane matching rules; however, they are less widely adopted and less representative across datasets compared to the standard F1 score. 
The F1 score is computed as:
\begin{equation}
    F_{1} = \frac{2 \times \text{Precision} \times \text{Recall}}{\text{Precision} + \text{Recall}},
\end{equation}
where Precision and Recall are defined based on the number of true positives (TP), false positives (FP), and false negatives (FN) as:
\[
\text{Precision} = \frac{TP}{TP + FP}, \quad \text{Recall} = \frac{TP}{TP + FN}.
\]

A predicted lane is considered a true positive if it sufficiently overlaps with a ground-truth lane. Specifically, each lane is projected onto the image plane, and the \gls{iou} between predicted and ground-truth lanes is computed. If the \gls{iou} exceeds a threshold (typically 0.5), the predicted lane is counted as a true positive. Otherwise, it is counted as a false positive, and unmatched ground-truth lanes are treated as false negatives.

\subsection{Performance Estimation Method Baselines}

In this section, we adapt five well-performing methods from image classification to lane detection tasks as our baselines. These methods are categorized into three types based on their underlying principles: confidence-based (\gls{ac}, \gls{doc}, \gls{atc}), feature distance-based (\gls{fid}), and energy-based (\gls{energy}) approaches.

Since the original methods were designed for image-level predictions, whereas lane detection outputs multiple lane instances per image, we adjust the unit of estimation from image-level to object-level. Specifically, all predicted lanes from each sample in a dataset are collected and treated individually for the purpose of performance estimation. 

In cases where no lanes are predicted across the entire dataset, the estimated performance score is set to zero.

\subsubsection{Average Confidence (AC)\cite{baselineood2016}}
One of the simplest methods is \gls{ac}, which calculates the average softmax confidence scores of all predicted lane instances to estimate performance. Formally, the AC score is computed as:

\begin{equation}
    \hat{F1} = AC(f, D^{t}) = \frac{\sum_{i=1}^{N} \sum_{j=1}^{M_i} c_{i,j}}
       {\sum_{i=1}^{N} M_i},
\end{equation}

where $N$ is the number of samples in $D^t$, \( M_i \) denotes the number of lanes predicted in sample \( x_i \), and \( c_{i,j} \) is the softmax score of the \( j \)-th predicted lane in sample \( i \).


\subsubsection{Difference of Confidence (DOC)\cite{doc2021}}

\gls{doc} was proposed to mitigate the overconfidence problem inherent in \gls{ac}. It calculates an offset based on the difference between the actual F1 score and the average confidence on the source domain, and applies this correction when estimating the target domain performance. A validation dataset from the source domain is required to compute this offset prior to deployment.

Following~\cite{doc2021}, we first compute the difference between the ground-truth F1 score and the average confidence (AC) on the source dataset \( D^s \). This offset is then added to the AC computed on the target dataset \( D^t \) to obtain the estimated F1 score. Formally, the procedure is expressed as:

\begin{equation}
    \text{offset} = F1^{v} - AC(f, D^v),
\end{equation}
\begin{equation}
    \hat{F1} = AC(f, D^t) + \text{offset},
\end{equation}

where \( F1^v \) denotes the actual F1 score of the model \( f \) on the validation dataset \( D^v \) from source domain.

\subsubsection{Average Threshold Confidence (ATC)\cite{atc2022}}

\gls{atc} is another method designed to address the overconfidence issue. It learns a confidence threshold \( t \) from validation datasets such that the proportion of predictions with confidence exceeding \( t \) aligns with the actual F1 score observed on the source domain.
Formally, the estimated F1 score on the target domain is computed as:

\begin{equation}
    \hat{F1} = ATC(f, D^{t}) = \mathbb{E}[\mathbb{I}[c_{i,j} > t]],
\end{equation}

where \(\mathbb{E}\) is the expectation function, \( \mathbb{I}[\cdot] \) is the indicator function, and \( c_{i,j} \) denotes the confidence of the \( j \)-th predicted lane in sample \( i \).

The threshold \( t \) is determined by solving an optimization problem on a collection of validation datasets \(\{D^{v,(k)}\}_{k=1}^{K}\), aiming to minimize the discrepancy between the estimated \(\hat{F1}\) and the actual F1 score on the validation data.

\subsubsection{Fréchet Inception Distance (FID)}

\gls{fid} is a commonly used metric in generation and domain adaptation tasks to quantify the similarity between two feature distributions. It is also used in~\cite{frechet2021} to quantify the domain shift between datasets and estimate classifier accuracy. In our setting, we adapt it for lane feature embeddings.
The FID score between two distributions is computed as:

\begin{equation}
    \text{dist}(D^{s}, D^{t}) = \|\mu_{s} - \mu_{t}\|_{2}^{2}  + \mathrm{Tr}\,\!\bigl(\Sigma_{s} + \Sigma_{t} - 2 (\Sigma_{s}\Sigma_{t})^{\tfrac{1}{2}}\bigr),
\end{equation}

where \(\mu\) and \(\Sigma\) are the mean and covariances of all lane feature embeddings extracted from the training dataset $D^s$ and the testing dataset $D_t$, respectively. $\mathrm{Tr}$ denotes the trace operator, i.e., the sum of the diagonal elements of a matrix.

The FID value ranges from zero (identical distributions) to arbitrarily large values (completely different distributions). Since the FID score does not naturally lie within the [0, 1] range like the F1 score, a regression model is employed to map the distribution distance to an estimated F1 score.
Formally, the estimation is expressed as:

\begin{equation}
    \hat{F1} = \text{Regression}\bigl(\text{dist}(D^s, D^t)\bigr),
\end{equation}

where the regression model is trained on a collection of validation datasets to fit the relationship between distribution distance and F1 score.



\subsubsection{Energy-based Model (EBM)}

Energy-based methods leverage the model's output logits to estimate the uncertainty of predictions without requiring explicit probability calibration. In our context, the energy score of a sample is used as a proxy to predict the model's F1 score on an unlabeled target dataset.

Following~\cite{energy2021}, we compute an energy score for each candidate lane based on its binary classification logits. Given a candidate lane \( l \), the energy is defined as:

\begin{equation}
    \text{Energy}(l, f) = -T \cdot \log \left( \sum_{j=1}^{2} \exp\left( \frac{f_j(l)}{T} \right) \right),
\end{equation}

where \( f_j(l) \) denotes the logit output by model \( f \) for the \( j \)-th class (lane / non-lane), and \( T \) is a temperature parameter calibrated on a validation set. Lower energy scores typically indicate more confident and accurate predictions, while higher energy scores suggest uncertainty and potential errors.

The energy scores of all lanes in dataset $D^t$ are aggregated together by average operation. Similar to \gls{fid}, a regression model is further trained to map the average energy to the estimated F1 score, as energy values are not naturally scaled to [0, 1].
Formally, the estimation process can be summarized as:

\begin{equation}
    \hat{F1} = \text{Regression}\left(\mathbb{E}\left[\text{Energy}(l, f)\right]_{l \in D^t}\right),
\end{equation}

where \(\mathbb{E}[\cdot]\) denotes the mean over all predicted lanes in the dataset. The regression model is fitted using a set of validation datasets, capturing the relationship between energy scores and actual F1 scores. In our experiments, we follow the setup of~\cite{frechet2021} and use a linear regression model for both \gls{fid} and \gls{energy}.

\begin{figure}[t]
    \centering
    \includegraphics[width=\linewidth]{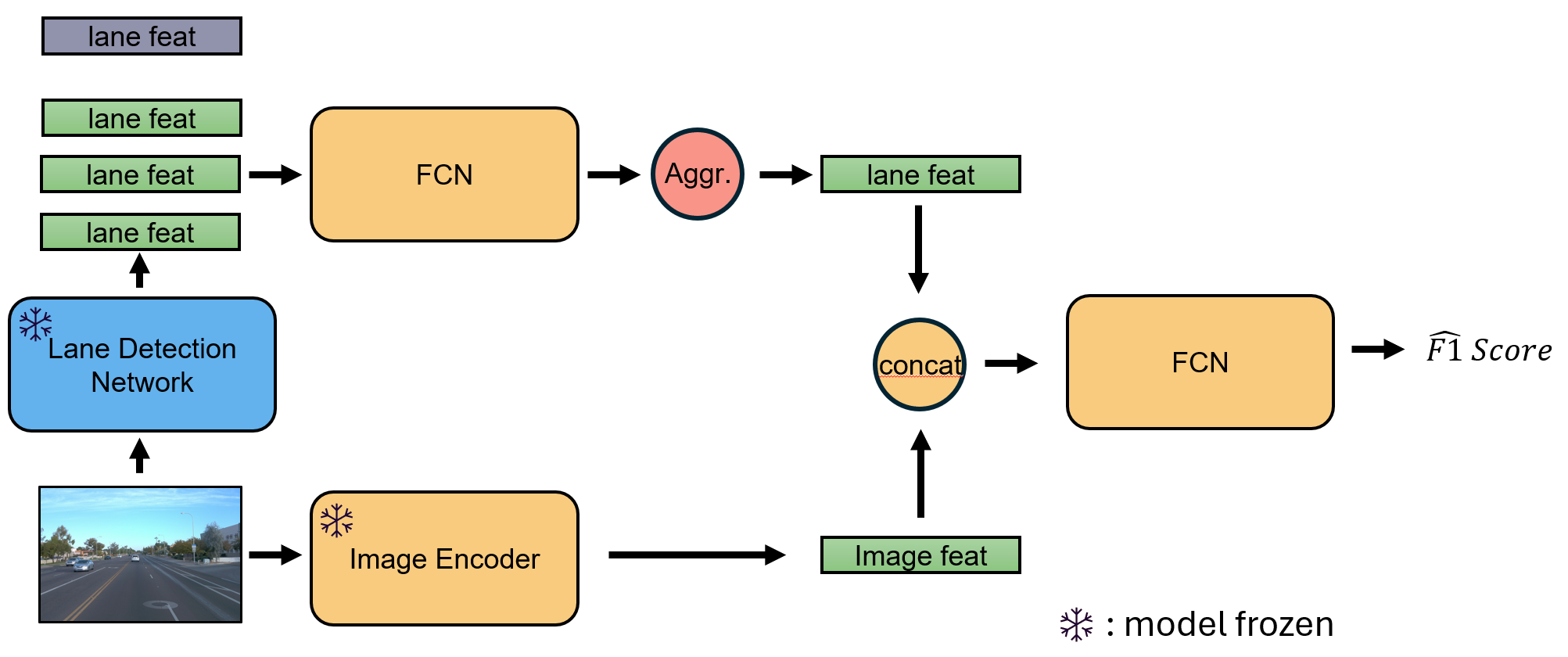}
    \caption{Our model predicts F1 score for lane detection by aggregating lane features via a fully connected layer and aggregation operator, concatenating them with image features from a pretrained image encoder, and applying a final regression head.}
    \label{fig:model}
\end{figure}

\section{Proposed Framework}

\subsection{Framework Design}

Although the aforementioned performance estimation methods have shown strong results in image classification, their effectiveness in lane detection remains unclear due to fundamental differences in task structure.
First, lane detection produces a variable number of outputs per image, whereas classification tasks yield exactly one class label per input. In particular, the case of zero predicted lanes introduces ambiguity: it may represent a correct outcome (e.g., no lanes present) or a failure (e.g., missed detections). Most baseline methods cannot handle this uncertainty. 

Second, the logits and feature representations extracted from the \gls{mut} may lose effectiveness under domain shift, as the network is exposed to unseen distributions.

To address these challenges, we propose \gls{laneperf}, a novel performance estimation framework tailored for lane detection. As illustrated in Figure~\ref{fig:model}, \gls{laneperf} extends traditional performance estimation by explicitly handling zero-lane outputs and incorporating both image-level and lane-level features, improving robustness across diverse driving scenarios.

To mitigate domain shift without introducing bias, we employ an image encoder to extract image-level features that capture general scene characteristics such as lighting, road conditions, and weather. The image encoder should be pretrained on a large-scale dataset  (e.g., ImageNet or a similar foundation dataset) to provide stronger domain-generalizable representations. For lane-level processing, we address the variable number of detected lanes using a DeepSets architecture~\cite{zaheer2017deep}, which processes each lane’s feature vector through a \gls{fcn} and aggregates them into a unified representation. In cases where the model outputs no detected lanes, the input set to the DeepSets module becomes empty. To handle this, we introduce a learnable default lane feature to be a placeholder, allowing the network to make meaningful predictions even in the absence of output from the \gls{mut}.

The image features and aggregated lane features are then concatenated and fed into a second \gls{fcn} to predict the sample’s performance score $F1$. This design ensures robust performance estimation, even in unseen challenging scenarios like rainy or night conditions.

\subsection{Model Details}

We select a pretrained CLIP image encoder (ViT-B/32)~\cite{radford2021learning} in our framework. It was trained on the ImageNet dataset~\cite{deng2009imagenet} containing 1.28 million images. The aggregation operator used in our framework is a mean pooling. Each \gls{fcn} consists of two layers, implemented as linear projections followed by ReLU activations. The final output layer is a sigmoid function, constraining the predicted score to the range \([0, 1]\).

\begin{table}[t]
\centering
\caption{F1 scores of CLRNet trained on different source domains and evaluated on various target domains. The domains are categorized according to the OpenLane dataset, grouped into scene, weather, and hours.}
\label{tab:dataset}
\begin{tabular}{llll}
\toprule
Domain & Source & Target & F1 Score \\
\midrule
\multirow{4}{*}{Scene} & \multirow{4}{*}{Residential Area} & Residential Area Val & 0.60 \\
 &  & Highway & 0.60 \\
 &  & Suburbs & 0.57 \\
 &  & Urban & 0.41 \\
 &  & Parking lot & 0.37 \\
\midrule
\multirow{4}{*}{Weather} & \multirow{4}{*}{Clear} & Clear Val & 0.61 \\
 &  & Cloudy & 0.55 \\
 &  & Rainy & 0.47 \\
 &  & Foggy & 0.63 \\
 &  & Overcast & 0.54 \\
\midrule
\multirow{2}{*}{Hours} & \multirow{2}{*}{Daytime} & Daytime Val & 0.63 \\
 &  & Night & 0.61 \\
 &  & Dawn/Dusk & 0.56 \\
\bottomrule
\end{tabular}
\end{table}

\section{Experiment Setup}
\label{sec:experiment}

This section introduces the details of our experiment setups.

\begin{table*}[htbp]
\centering
\caption{MAE and Spearman $\rho$ for six performance-estimation methods (\gls{laneperf}, \gls{ac}, \gls{doc}, \gls{atc}, \gls{fid}, \gls{energy}) applied to CLRNet on OpenLane. CLRNet is trained on each source domain and evaluated on the another target domain; lower MAE and higher $\rho$ indicate better effectiveness.}
\label{tab:main}
\setlength{\tabcolsep}{2pt}
\renewcommand{\arraystretch}{1.15}
\newcolumntype{Y}{>{\centering\arraybackslash}X}

\begin{tabularx}{\textwidth}{l l *{12}{Y}}
\toprule
\multicolumn{2}{c}{} & \multicolumn{6}{c}{\textbf{MAE} $\downarrow$} & \multicolumn{6}{c}{\boldmath{$\rho$} $\uparrow$} \\
\cmidrule(lr){3-8}\cmidrule(lr){9-14}
& & \makecell{LanePerf\\(Ours)} & AC & DOC & ATC & FID  & EMB & \makecell{LanePerf\\(Ours)} & AC & DOC & ATC & FID  & \ EMB \\
\midrule
\multirow{4}{*}{Residential}
& Highway & 0.103 & \textbf{0.074} & 0.174 & 0.145 & 0.079 & 0.223 & 0.591 & 0.713 & 0.713 & \textbf{0.726} & 0.592 & 0.625 \\
& Urban & \textbf{0.119} & 0.135 & 0.238 & 0.155 & 0.206 & 0.167 & \textbf{0.721} & 0.655 & 0.655 & 0.593 & 0.625 & 0.534 \\
& Suburbs & \textbf{0.102} & 0.122 & 0.187 & 0.154 & 0.142 & 0.206 & \textbf{0.779} & 0.673 & 0.673 & 0.750 & 0.662 & 0.696 \\
& Parking & \textbf{0.188} & 0.195 & 0.231 & 0.225 & 0.249 & 0.263 & 0.511 & 0.418 & 0.418 & \textbf{0.547} & 0.333 & 0.169 \\
\cmidrule(lr){2-14}
& Scene Avg. & \textbf{0.128} & 0.132 & 0.207 & 0.170 & 0.169 & 0.215 & 0.651 & 0.615 & 0.615 & \textbf{0.654} & 0.553 & 0.506 \\
\midrule
\multirow{4}{*}{Clear}
& Cloudy & \textbf{0.119} & 0.145 & 0.157 & 0.203 & 0.155 & 0.226 & \textbf{0.723} & 0.618 & 0.618 & 0.631 & 0.652 & 0.601 \\
& Overcast & \textbf{0.107} & 0.128 & 0.159 & 0.191 & 0.155 & 0.222 & \textbf{0.830} & 0.706 & 0.706 & 0.694 & 0.733 & 0.696 \\
& Foggy & \textbf{0.087} & 0.120 & 0.126 & 0.203 & 0.146 & 0.239 & \textbf{0.764} & 0.732 & 0.732 & 0.540 & 0.764 & 0.663 \\
& Rainy & \textbf{0.117} & 0.135 & 0.171 & 0.241 & 0.180 & 0.180 & \textbf{0.740} & 0.642 & 0.642 & 0.385 & 0.605 & 0.446 \\
\cmidrule(lr){2-14}
& Weather Avg. & \textbf{0.108} & 0.132 & 0.153 & 0.210 & 0.159 & 0.217 & \textbf{0.764} & 0.674 & 0.674 & 0.562 & 0.689 & 0.602 \\
\midrule
\multirow{2}{*}{Hours}
& Dawn/Dusk & \textbf{0.118} & 0.145 & 0.162 & 0.172 & 0.176 & 0.239 & \textbf{0.804} & 0.742 & 0.742 & 0.710 & 0.710 & 0.513 \\
& Night & 0.113 & \textbf{0.098} & 0.150 & 0.143 & 0.190 & 0.162 & \textbf{0.807} & 0.793 & 0.793 & 0.770 & 0.607 & 0.649 \\
\cmidrule(lr){2-14}
& Hours Avg. & \textbf{0.116} & 0.122 & 0.156 & 0.158 & 0.183 & 0.201 & \textbf{0.806} & 0.767 & 0.767 & 0.740 & 0.659 & 0.581 \\
\midrule
All Avg. & & \textbf{0.117} & 0.130 & 0.176 & 0.183 & 0.168 & 0.213 & \textbf{0.727} & 0.669 & 0.669 & 0.635 & 0.628 & 0.559 \\
\bottomrule
\end{tabularx}
\end{table*}

\textbf{Cross-Domain Setup}. Lane detection across diverse domains remains a significant challenge, yet few publicly available datasets are specifically designed to address this issue. For example, widely used 2D lane detection datasets such as CULane~\cite{scnn2018}, LLAMAS~\cite{llamas2019}, and TuSimple~\cite{tusimple2020} lack domain-specific annotations, limiting their utility for cross-domain evaluation.
We initially explored a cross-dataset setup to simulate domain shifts; however, differences in annotation styles across datasets hindered consistent comparisons under a unified framework. Consequently, we selected the OpenLane dataset~\cite{chen2022persformer} for our study, which provides diverse domain labels enabling structured evaluation.
The domain shifts are categorized into three dimensions based on its annotations:

\begin{itemize}
    \item Scenes: describe different driving environments, including residential areas, highways, suburbs, urban environments, and parking lots.
    \item Weather: describe different weather conditions, including clear, cloudy, rainy, foggy, and overcast.
    \item Hours: describe different time of the day, including daytime, nighttime, and dawn/dusk.
\end{itemize}

To ensure sufficient training data, we select the subset with the largest number of frames as the source domain. The remaining subsets served as target domains and were excluded from both the training of the lane detection model and the training of the performance estimation model.
Additionally, in Table \ref{tab:dataset}, we report the actual F1 scores for each target domain to provide insight into the relative difficulty of the \gls{mut} across different domain shifts.
For both the performance estimation model fitting and inference stages, multiple datasets are required for training and evaluation. To this end, we follow the OpenLane data format and treat each segment as an individual mini-dataset. Each segment consists of approximately 200 consecutive frames, representing a continuous video recording of a driving scenario.
A significant performance drop within a mini-dataset can indicate a potential corner case, highlighting scenarios that pose particular challenges to the reliability of \gls{adas} systems.

\textbf{Metrics}. To evaluate the effectiveness of performance estimation methods, we report two metrics by comparing the estimated $\hat{F1}$ score and the actual $F1$ score: 
\begin{itemize}
    \item Mean Absolute Error (MAE), which represents the error between the predicted score and the actual score. Ranges from 0 to 1, where 0 is no error at all, 1 is complete error:
    \begin{equation}
        \text{MAE} = \frac{1}{N} \sum_{i=1}^{N} \left| F1_{i} - \hat{F1}_{i} \right|.
        \label{eq:mae}
    \end{equation}
    
    \item Spearman's rank correlation coefficient ($\rho$), which measures the monotonic relationship between the predicted and actual $F1$ scores. Instead of using the coefficient of determination ($R^2$), we adopt \gls{rho}, as the relationship between predicted and actual performance may not be strictly linear. A monotonic correlation is therefore more appropriate for this task. The value of $\rho$ ranges from -1 to 1, where -1 indicates a perfect negative correlation, 1 indicates a perfect positive correlation, and 0 indicates no correlation:

    \begin{equation}
        \rho = 1 - \frac{6 \sum_{i=1}^N d_i^2}{N(N^2 - 1)}.
        \label{eq:spearman}
    \end{equation}
\end{itemize}

\textbf{Lane Detection Model.} We employ CLRNet~\cite{clrnet2022}, which achieves state-of-the-art performance and ranks among the top on multiple dataset benchmarks~\cite{scnn2018, llamas2019, tusimple2020}. The model is trained separately on the training splits of three datasets (scene: residential area, weather: clear, hours: daytime), with all hyperparameters kept consistent. The raw output from CLRNet is post-processed using \gls{nms} and a predefined confidence threshold. Following the original implementation, the confidence threshold is set to 0.4. Only the lane after the confidence threshold and \gls{nms} are kept, with its lane feature vector before head, logits value and softmax scores stored.
\section{Evaluation}
\label{sec:evaluation}

In this section, we present the empirical results of five baseline methods and our proposed \gls{laneperf} under three types of domain shift in the OpenLane benchmark. The main results are summarized in Table~\ref{tab:main}, which reports both the average \gls{mae} and \gls{rho} across scene, weather, and hours domains, as well as the overall average. We begin by evaluating the five baseline methods, followed by a detailed analysis of our \gls{laneperf}. Finally, we include two ablation studies to validate the contribution of each component in our design.

\subsection{Evaluation of the Baseline Methods}
\label{section:eval:baseline}

Among the five baseline methods, \gls{ac} achieves the best performance overall across the baselines, despite being one of the earliest and simplest methods.
Compared to \gls{ac}, the other baselines, which were originally proposed to address softmax overconfidence, are less effective. As the validation datasets are from the same domain of the training data, the calibration process tends to overfit the models, thereby reducing generalization to unseen domains. 
Notably, all methods degrade significantly in the Parking lot scenario, but \gls{energy} suffers the most, with a $\rho$ as low as 0.169, suggesting that this method is particularly sensitive to out-of-distribution inputs. The distance-based method \gls{fid} falls between confidence-based and density-based approaches, likely because lane features lack sufficient fine-grained structural information.
\begin{figure}[t]
    \centering
    \includegraphics[width=\linewidth]{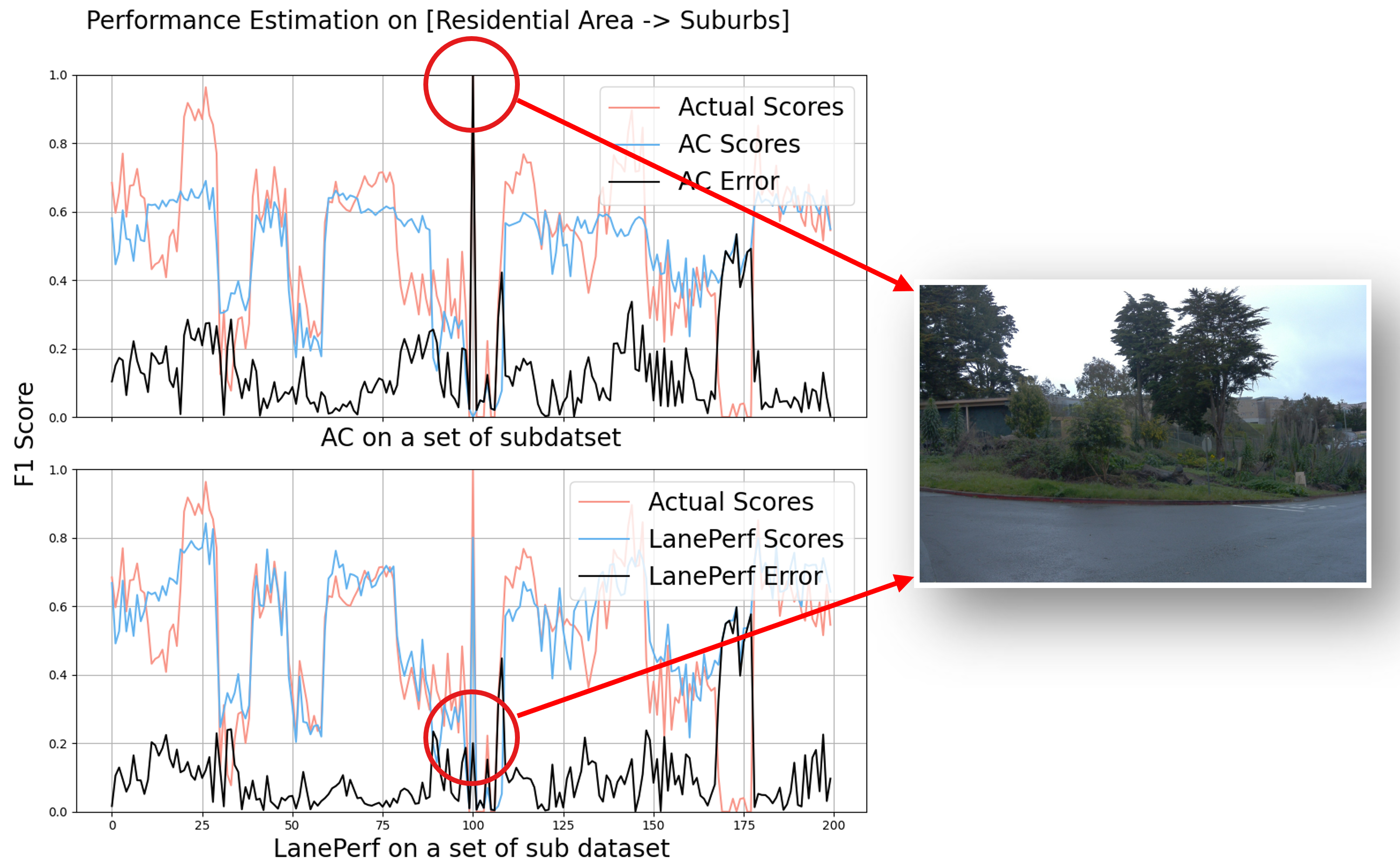}
    \caption{An example of non-lane cases. Blue: estimated score, red: actual score, black: absolute error. Each point on the x-axis represents a mini-dataset consisting of 50 images. We show the first 200 subsets from the Suburbs split.}
    \label{fig:seq_zero}
\end{figure}

When comparing results across different domain shifts, scene changes pose the greatest challenge for performance estimation methods, yielding the highest average \gls{mae} and the lowest \gls{rho}. Weather variations have the least impact, while time-of-day shifts (hours) fall in between. These results suggest that lane detection is more sensitive to scene layout and illumination conditions, making them the most challenging factors for performance estimation.

\subsection{Evaluation of \gls{laneperf}}

\gls{laneperf} achieves the best overall performance, with the lowest \gls{mae} (0.117) and the highest Spearman’s \gls{rho} (0.727) across all transfer settings. Compared to the strongest baseline, \gls{ac} (\gls{mae} 0.130, $\rho$ 0.669), \gls{laneperf} reduces estimation error by approximately 10\% and improves rank correlation by 5.8\%. 
It ranks first in both MAE and $\rho$ on 8 out of 10 cross-domain evaluations, whereas the best-performing baselines—\gls{ac} and \gls{atc}—only achieve top results in two domains each. Notably, \gls{laneperf} maintains strong performance under adverse conditions such as poor weather (all four weather types) and significant scene variations (urban, suburbs), demonstrating its ability to capture domain shifts through semantically meaningful image features. In the night setting, although \gls{laneperf} has a slightly higher \gls{mae} than \gls{ac}, likely due to the difficulty of extracting semantic features under low illumination, it still achieves a higher \gls{rho}, indicating more reliable performance ranking across models.

\subsection{LanePerf on Zero-Lane Situation}

To evaluate the effectiveness of \gls{laneperf} in non-lane cases, we reduce the dataset size from 200 to 50 samples, increasing the likelihood of encountering such cases. Figure~\ref{fig:seq_zero} illustrates an example from the [Residential Area $\rightarrow$ Suburbs] setup. During a turning maneuver, both the predictions and ground-truth labels contain no lanes. In this case, \gls{ac} defaults to a score of 0, as it lacks any input features to process. In contrast, \gls{laneperf} successfully leverages image-level information to produce a score closer to the actual value, which demonstrates that our proposed method remains robust even when explicit lane output is entirely absent.


\subsection{Ablation Study on Image and Lane Features}

\begin{figure}[t]
    \centering
    \includegraphics[width=\linewidth]{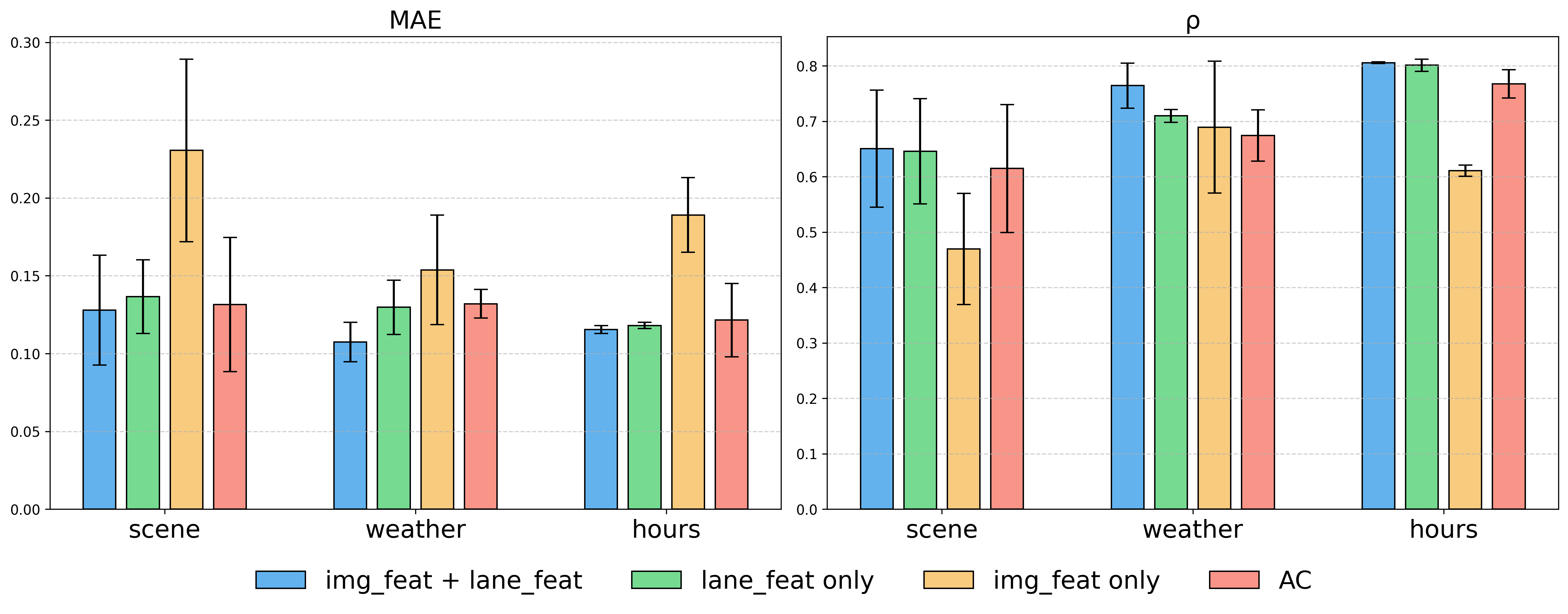}
    \caption{Ablation study on the influence of using image features and lane feature vectors. The plot shows the mean and standard deviation of "scene", "weather", and "hours" domain shift.}
    \label{fig:ablation_feat}
\end{figure}

To quantify the contribution of image embeddings, we carried out an ablation study with four configurations of \gls{laneperf} (For overview of \gls{laneperf} compare Figure~\ref{fig:model}): (i) $img\_feat + lane\_feat$, (ii) $lane\_feat\ only$, (iii) $img\_feat\ only$, and (iv) the strongest baseline, \gls{ac}. The results are shown in Figure \ref{fig:ablation_feat}.
The $img\_feat + lane\_feat$ variant delivers the lowest and most consistent \gls{mae} on all three types of domain shift. Its \gls{rho} values are also the highest, with much lower standard deviation compared to $lane\_feat\ only$ and \gls{ac}. For domain type "hours", $lane\_feat\ only$ shows competitive effectiveness as $img\_feat + lane\_feat$, which suggests that image embeddings are less informative for illuminations.
At the other extreme, $img\_feat\ only$ performs worst, confirming that discarding \gls{mut} information is not reliable.

\subsection{Ablation Study on Pretrained Image Encoder}

\begin{figure}[t]
    \centering
    \includegraphics[width=\linewidth]{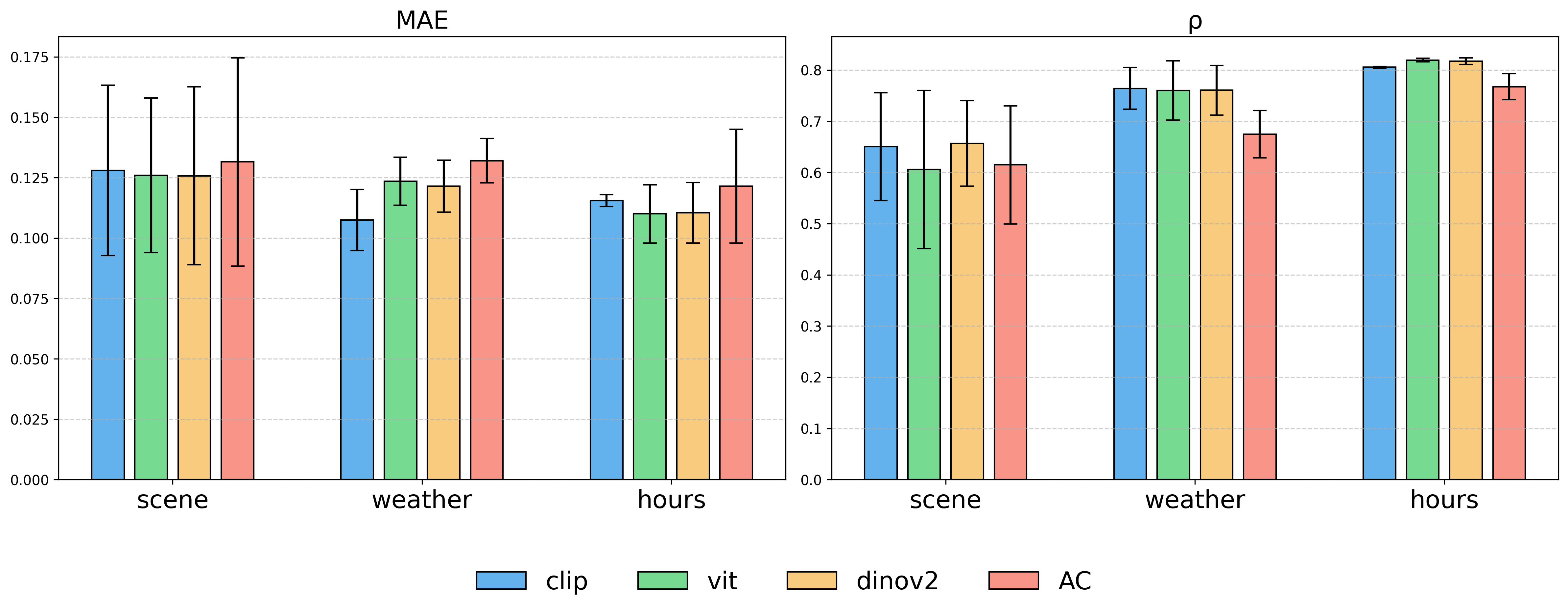}
    \caption{Ablation study on the selection of pretrained image encoder. The plot shows the mean and standard deviation across all domain-shift combinations.}
    \label{fig:ablation_model}
\end{figure}

Our framework supports the flexible use of different pretrained image encoders. In Figure \ref{fig:ablation_model}, we compare two additional popular foundation models alongside CLIP: ViT (vit-base-patch16-224)~\cite{dosovitskiy2020image} and DINOv2 (dinov2-base)~\cite{oquab2023dinov2}.
In the “scene” domain, all three encoders achieve nearly identical \gls{mae}. CLIP attains the lowest \gls{mae} in the “weather” domain but incurs a marginally higher \gls{mae} in the “hours” domain compared to ViT and DINOv2. Across all domains, each pretrained encoder exhibits substantially lower MAE standard deviation than the baseline \gls{ac} method, most notably under the domain shift of “hours”. For \gls{rho}, ViT slightly lags behind the other three methods in the "scene" domain but remains effective in both "weather" and "hours".

\section{Conclusion}
\label{sec:conclusion}

In this paper, we addressed the largely unexplored problem of \textit{label-free} performance estimation for lane detection under domain shift. 
We adapted and analyzed five well-performing performance estimation methods from image classification to lane detection as our baseline.
Based on these insights, we introduced \gls{laneperf}, a lightweight framework that leverages a powerful pretrained image encoder and lane features.
Extensive experiments across various real-world domain shifts demonstrate our \gls{laneperf} achieved state-of-the-art results, outperforming all baselines. Our method enables fast and scalable performance estimation without costly annotations, accelerating dataset evaluation and model development cycles.

We hope our work lays a foundation for future research in cross-domain model testing and label-free performance evaluation for \gls{adas}.

{
    \bibliographystyle{IEEEtran}
    \small
    \bibliography{references}

\begin{thebibliography}{10}
\providecommand{\url}[1]{#1}
\csname url@samestyle\endcsname
\providecommand{\newblock}{\relax}
\providecommand{\bibinfo}[2]{#2}
\providecommand{\BIBentrySTDinterwordspacing}{\spaceskip=0pt\relax}
\providecommand{\BIBentryALTinterwordstretchfactor}{4}
\providecommand{\BIBentryALTinterwordspacing}{\spaceskip=\fontdimen2\font plus
\BIBentryALTinterwordstretchfactor\fontdimen3\font minus \fontdimen4\font\relax}
\providecommand{\BIBforeignlanguage}[2]{{%
\expandafter\ifx\csname l@#1\endcsname\relax
\typeout{** WARNING: IEEEtran.bst: No hyphenation pattern has been}%
\typeout{** loaded for the language `#1'. Using the pattern for}%
\typeout{** the default language instead.}%
\else
\language=\csname l@#1\endcsname
\fi
#2}}
\providecommand{\BIBdecl}{\relax}
\BIBdecl

\bibitem{lanenet2018}
Z.~Wang, W.~Ren, and Q.~Qiu, ``Lanenet: Real-time lane detection networks for autonomous driving,'' \emph{arXiv preprint arXiv:1807.01726}, 2018.

\bibitem{scnn2018}
X.~Pan, J.~Shi, P.~Luo, X.~Wang, and X.~Tang, ``Spatial as deep: Spatial cnn for traffic scene understanding,'' in \emph{Proceedings of the AAAI conference on artificial intelligence}, 2018.

\bibitem{lineCNN2020}
X.~Li, J.~Li, X.~Hu, and J.~Yang, ``Line-cnn: End-to-end traffic line detection with line proposal unit,'' \emph{IEEE Transactions on Intelligent Transportation Systems}, 2019.

\bibitem{laneatt2021}
L.~Tabelini, R.~Berriel, T.~M. Paixao, C.~Badue, A.~F. De~Souza, and T.~Oliveira-Santos, ``Keep your eyes on the lane: Real-time attention-guided lane detection,'' in \emph{Proceedings of the IEEE/CVF conference on computer vision and pattern recognition}, 2021.

\bibitem{polylanenet2021}
------, ``Polylanenet: Lane estimation via deep polynomial regression,'' in \emph{2020 25th International Conference on Pattern Recognition (ICPR)}, 2021.

\bibitem{clrnet2022}
T.~Zheng, Y.~Huang, Y.~Liu, W.~Tang, Z.~Yang, D.~Cai, and X.~He, ``Clrnet: Cross layer refinement network for lane detection,'' in \emph{Proceedings of the IEEE/CVF conference on computer vision and pattern recognition}, 2022.

\bibitem{domainshift2009}
J.~Qui{\~n}onero-Candela, M.~Sugiyama, A.~Schwaighofer, and N.~D. Lawrence, \emph{Dataset shift in machine learning}.\hskip 1em plus 0.5em minus 0.4em\relax Mit Press, 2009.

\bibitem{frechet2021}
W.~Deng and L.~Zheng, ``Are labels always necessary for classifier accuracy evaluation?'' in \emph{Proceedings of the IEEE/CVF conference on computer vision and pattern recognition}, 2021.

\bibitem{baselineood2016}
D.~Hendrycks and K.~Gimpel, ``A baseline for detecting misclassified and out-of-distribution examples in neural networks,'' \emph{arXiv preprint arXiv:1610.02136}, 2016.

\bibitem{atc2022}
S.~Garg, S.~Balakrishnan, Z.~C. Lipton, B.~Neyshabur, and H.~Sedghi, ``Leveraging unlabeled data to predict out-of-distribution performance,'' \emph{arXiv preprint arXiv:2201.04234}, 2022.

\bibitem{doc2021}
D.~Guillory, V.~Shankar, S.~Ebrahimi, T.~Darrell, and L.~Schmidt, ``Predicting with confidence on unseen distributions,'' in \emph{Proceedings of the IEEE/CVF international conference on computer vision}, 2021.

\bibitem{maha2018}
K.~Lee, K.~Lee, H.~Lee, and J.~Shin, ``A simple unified framework for detecting out-of-distribution samples and adversarial attacks,'' \emph{Advances in neural information processing systems}, 2018.

\bibitem{knn2022}
Y.~Sun, Y.~Ming, X.~Zhu, and Y.~Li, ``Out-of-distribution detection with deep nearest neighbors,'' in \emph{International Conference on Machine Learning}, 2022.

\bibitem{energy2021}
W.~Liu, X.~Wang, J.~Owens, and Y.~Li, ``Energy-based out-of-distribution detection,'' \emph{Advances in neural information processing systems}, 2020.

\bibitem{energy2022}
D.~Duvenaud, J.~Wang, J.~Jacobsen, K.~Swersky, M.~Norouzi, and W.~Grathwohl, ``Your classifier is secretly an energy based model and you should treat it like one,'' in \emph{ICLR 2020}, 2020.

\bibitem{lanedetsurvey2024}
X.~He, H.~Guo, K.~Zhu, B.~Zhu, X.~Zhao, J.~Fang, and J.~Wang, ``Monocular lane detection based on deep learning: A survey,'' \emph{arXiv preprint arXiv:2411.16316}, 2024.

\bibitem{lanedettranssurvey2025}
J.~Bi, Y.~Song, Y.~Jiang, L.~Sun, X.~Wang, Z.~Liu, J.~Xu, S.~Quan, Z.~Dai, and W.~Yan, ``Lane detection for autonomous driving: Comprehensive reviews, current challenges, and future predictions,'' \emph{IEEE Transactions on Intelligent Transportation Systems}, 2025.

\bibitem{llamas2019}
K.~Behrendt and R.~Soussan, ``Unsupervised labeled lane markers using maps,'' in \emph{Proceedings of the IEEE/CVF international conference on computer vision workshops}, 2019.

\bibitem{tusimple2020}
TuSimple, ``Tusimple benchmark,'' \url{https://github.com/TuSimple/tusimple-benchmark/}, accessed: September 2020.

\bibitem{bridelanedetDA2020}
T.~Wen, D.~Yang, K.~Jiang, C.~Yu, J.~Lin, B.~Wijaya, and X.~Jiao, ``Bridging the gap of lane detection performance between different datasets: Unified viewpoint transformation,'' \emph{IEEE Transactions on Intelligent Transportation Systems}, 2020.

\bibitem{weaklyDA2024}
J.~Zhou, C.~Zhang, and J.~Beyerer, ``Towards weakly-supervised domain adaptation for lane detection,'' in \emph{Proceedings of the IEEE/CVF Conference on Computer Vision and Pattern Recognition}, 2024.

\bibitem{tempscaling2017}
C.~Guo, G.~Pleiss, Y.~Sun, and K.~Q. Weinberger, ``On calibration of modern neural networks,'' in \emph{International conference on machine learning}, 2017.

\bibitem{agreement2022}
C.~Baek, Y.~Jiang, A.~Raghunathan, and J.~Z. Kolter, ``Agreement-on-the-line: Predicting the performance of neural networks under distribution shift,'' \emph{Advances in Neural Information Processing Systems}, 2022.

\bibitem{perfpred_recon_seg2022}
A.~B{\"a}r, M.~Klingner, J.~L{\"o}hdefink, F.~H{\"u}ger, P.~Schlicht, and T.~Fingscheidt, ``Performance prediction for semantic segmentation by a self-supervised image reconstruction decoder,'' in \emph{Proceedings of the IEEE/CVF Conference on Computer Vision and Pattern Recognition}, 2022.

\bibitem{perfpred_seg2021}
M.~Klingner, A.~Bar, M.~Mross, and T.~Fingscheidt, ``Improving online performance prediction for semantic segmentation,'' in \emph{Proceedings of the IEEE/CVF Conference on Computer Vision and Pattern Recognition}, 2021.

\bibitem{zaheer2017deep}
M.~Zaheer, S.~Kottur, S.~Ravanbakhsh, B.~Poczos, R.~R. Salakhutdinov, and A.~J. Smola, ``Deep sets,'' \emph{Advances in neural information processing systems}, vol.~30, 2017.

\bibitem{radford2021learning}
A.~Radford, J.~W. Kim, C.~Hallacy, A.~Ramesh, G.~Goh, S.~Agarwal, G.~Sastry, A.~Askell, P.~Mishkin, J.~Clark \emph{et~al.}, ``Learning transferable visual models from natural language supervision,'' in \emph{International conference on machine learning}.\hskip 1em plus 0.5em minus 0.4em\relax PmLR, 2021, pp. 8748--8763.

\bibitem{deng2009imagenet}
J.~Deng, W.~Dong, R.~Socher, L.-J. Li, K.~Li, and L.~Fei-Fei, ``Imagenet: A large-scale hierarchical image database,'' in \emph{2009 IEEE conference on computer vision and pattern recognition}.\hskip 1em plus 0.5em minus 0.4em\relax Ieee, 2009, pp. 248--255.

\bibitem{chen2022persformer}
L.~Chen, C.~Sima, Y.~Li, Z.~Zheng, J.~Xu, X.~Geng, H.~Li, C.~He, J.~Shi, Y.~Qiao \emph{et~al.}, ``Persformer: 3d lane detection via perspective transformer and the openlane benchmark,'' in \emph{European Conference on Computer Vision}.\hskip 1em plus 0.5em minus 0.4em\relax Springer, 2022, pp. 550--567.

\bibitem{dosovitskiy2020image}
A.~Dosovitskiy, L.~Beyer, A.~Kolesnikov, D.~Weissenborn, X.~Zhai, T.~Unterthiner, M.~Dehghani, M.~Minderer, G.~Heigold, S.~Gelly \emph{et~al.}, ``An image is worth 16x16 words: Transformers for image recognition at scale,'' \emph{arXiv preprint arXiv:2010.11929}, 2020.

\bibitem{oquab2023dinov2}
M.~Oquab, T.~Darcet, T.~Moutakanni, H.~Vo, M.~Szafraniec, V.~Khalidov, P.~Fernandez, D.~Haziza, F.~Massa, A.~El-Nouby \emph{et~al.}, ``Dinov2: Learning robust visual features without supervision,'' \emph{arXiv preprint arXiv:2304.07193}, 2023.

\end{thebibliography}
}

\end{document}